\pdfoutput=1

\documentclass[11pt]{article}

\usepackage[]{EMNLP2023}
\usepackage{times}
\usepackage{latexsym}
\usepackage{booktabs}
\usepackage{multirow}
\usepackage{amsmath}
\usepackage{algorithm}
\usepackage{comment}
\usepackage{tabularray}
\usepackage{algpseudocode}
\usepackage[symbol]{footmisc}
\usepackage[T1]{fontenc}

\usepackage[utf8]{inputenc}

\usepackage{microtype}

\usepackage{inconsolata}

\usepackage{graphicx}

\title{Redefining Proactivity for Information Seeking Dialogue}

\author{Jing Yang Lee\textsuperscript{1}\footnotemark[1], Seokhwan Kim\textsuperscript{2}, Kartik Mehta\textsuperscript{3}, Jiun-Yu Kao\textsuperscript{3}, Yu-Hsiang Lin\textsuperscript{3,4}, Arpit Gupta\textsuperscript{3}\\
  Nanyang Technological University\textsuperscript{1}, Google Cloud AI\textsuperscript{2}, Amazon AGI\textsuperscript{3}\\
  jingyang001@e.ntu.edu.sg\textsuperscript{1}, seokhwankim@google.com\textsuperscript{2}\\\{kartim, jiunyk, guparpit\}@amazon.com\textsuperscript{3}, hitr2997925@gmail.com\textsuperscript{4}}

\begin{document}
\maketitle
\footnotetext[1]{Work done during internship at Amazon AGI.}
\begin{abstract}

Information-Seeking Dialogue (ISD) agents aim to provide accurate responses to user queries. While proficient in directly addressing user queries, these agents, as well as LLMs in general, predominantly exhibit reactive behavior, lacking the ability to generate proactive responses that actively engage users in sustained conversations. However, existing definitions of proactive dialogue in this context do not focus on how each response actively engages the user and sustains the conversation. Hence, we present a new definition of proactivity that focuses on enhancing the `proactiveness' of each generated response via the introduction of new information related to the initial query. To this end, we construct a proactive dialogue dataset comprising 2,000 single-turn conversations, and introduce several automatic metrics to evaluate response `proactiveness' which achieved high correlation with human annotation. Additionally, we introduce two innovative Chain-of-Thought (CoT) prompts, the 3-step CoT and the 3-in-1 CoT prompts, which consistently outperform standard prompts by up to 90\% in the zero-shot setting.
\end{abstract}

\section{Introduction}
\label{sec:intro}

\begin{figure}
    \centering
    \includegraphics[scale=0.21]{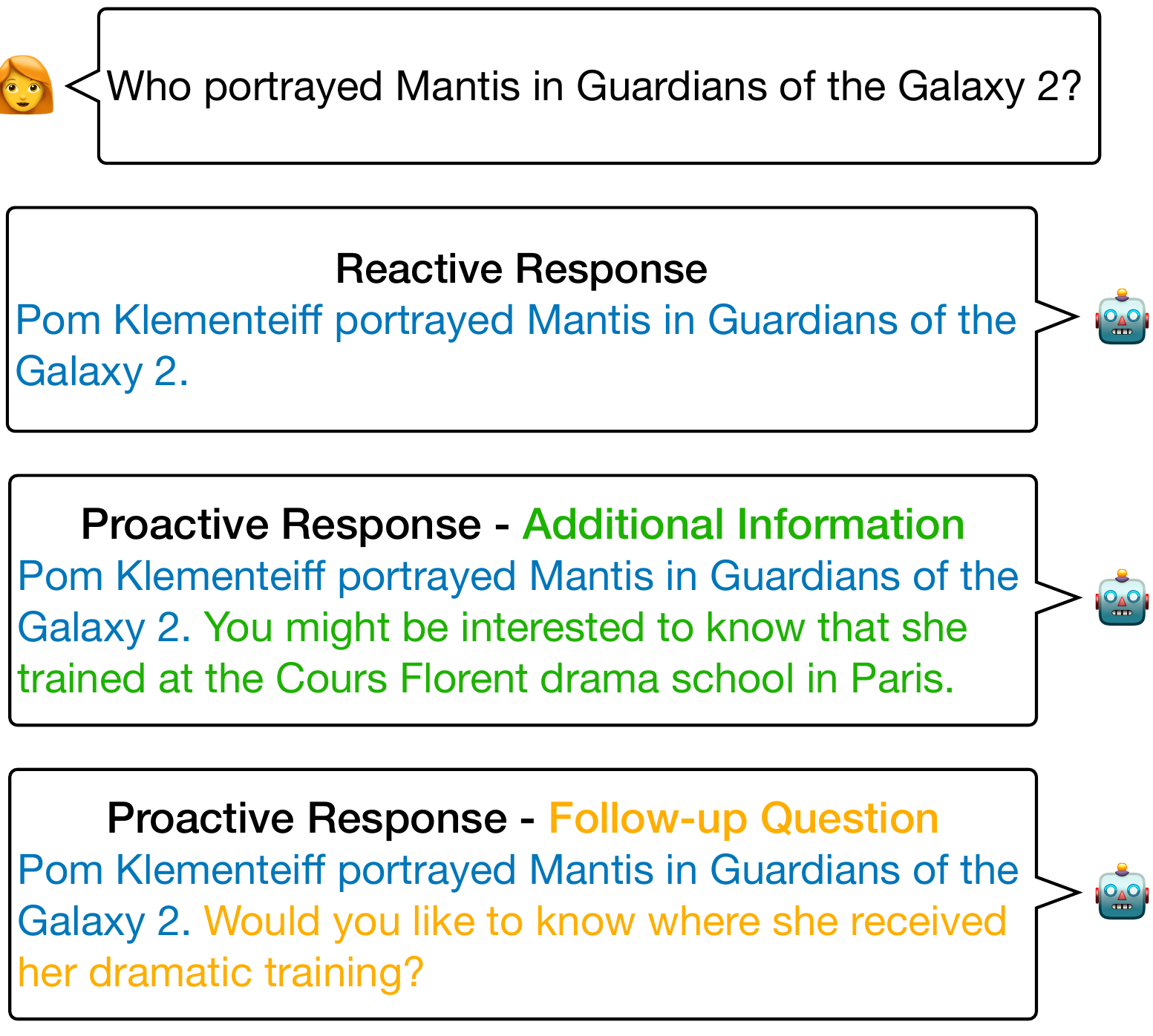}
    \caption{Illustration of a single user query and the corresponding reactive and proactive responses. Each proactive response corresponds to a specific proactive element type. The follow-up question is marked in \textcolor{orange}{orange text}, additional information is denoted by \textcolor{green}{green text}, and the answer component is indicated in \textcolor{blue}{blue text}.}
    \label{fig:dial_example}
\end{figure}

Generally, the aim of Information-Seeking Dialogue (ISD) agents \cite{dziri2022faithdial, nakamura-etal-2022-hybridialogue} is to generate an informative response which answers the user's query. In these interactions, users typically pose questions to obtain specific pieces of information, and the dialogue agent generates coherent responses which contains the information requested by the user. 
In recent years, Large Language Models (LLMs) have generally succeeded at achieving this goal \cite{Li_2023, braunschweiler2023evaluating}. However, current ISD agents, as well as LLMs in general, tend to be more reactive than proactive. An example of a reactive response is provided in Figure~\ref{fig:dial_example}. Responses generated by a reactive ISD agent would adequately address the user's query but fail to proactively engage the user. Once the requested information is provided, the conversation with the ISD agent naturally concludes.

In ISD, existing work on proactivity primarily focuses on generating clarifying questions and eliciting user preferences \cite{deng2023survey}, aiming to resolve ambiguity in the user's query or uncover their preference respectively. Current definitions of proactivity in ISD do not emphasize engaging the user or sustaining the conversation once the desired information has been provided. 
Hence, we introduce a novel definition of ISD proactivity that emphasizes generating responses that aim to sustain the interaction by proactively engaging the user via the introduction of new information \emph{pertinent to the initial query}. By proactively providing new related information, the agent can stimulate the user's interest, prompting further inquiries and sustaining the conversation. Hence, our definition of ISD proactivity focuses on actively delivering information related to the initial query in a conversational manner, thereby naturally guiding the conversation towards addressing multiple pieces of information, improving the overall informativeness during interactions with users and further enhancing user satisfaction~ \citep{deng2023survey, 10.1145/3234149}. Unlike prior definitions, we focus on the proactiveness of each individual response, evaluating them individually rather than as part of the entire conversation. This allows us to evaluate responses on specific criterion (Section 3).


According to our definition, a proactive response consists of the answer to the user's query and a proactive element, which refers to new information related to the initial query. The proactive element can be further categorized as either a Follow-up Question (FQ) or Additional Information(AI). Samples of proactive responses according to our definitions are also provided in Figure \ref{fig:dial_example}. 
It's important to note that this work does not encompass factual accuracy or information correctness. The focus is purely on syntactic and semantic proactivity. 

In this paper, our contributions are as follows:
\begin{enumerate}
    \item We introduce a novel response-level definition of proactivity for ISD.
    \item We construct a proactive dialogue corpus consisting of 2,000 single-turn conversations.
    \item We introduce a set of automatic metrics designed to measure the level of `proactiveness' in a response, according to our definition of proactive dialogue. Our metrics demonstrate high correlation with human annotation. 
    \item We propose two in-context Chain-of-Thought (CoT) prompts, namely the 3-step CoT prompt and the 3-in-1 CoT prompt, which outperform standard few-shot prompting. Additionally, utilizing our corpus, we demonstrate the efficacy of instruction-tuning in the context of proactive response generation.
    \item We demonstrate the efficacy of our approach in sustaining user interaction and improving conversational informativeness and in the multi-turn scenarios.
\end{enumerate}

\section{Related Work}

\textbf{Proactive Dialogue} Proactive dialogue encompasses various techniques for engaging users by steering conversations in specific directions. In the context of Open-Domain (OD) dialogue, some popular proactive dialogue tasks include: target-guided dialogue, prosocial dialogue, and non-collaborative dialogue. Target guided dialogue focuses on directing interactions toward predefined topics or entities, using methods such as response planning \cite{kishinami-etal-2022-target}, event-based knowledge graphs \cite{10.5555/3491440.3491985}, and commonsense bridging \cite{gupta-etal-2022-target}. Prosocial dialogue involves generating non-offensive responses that adhere to societal norms \cite{kim2022prosocialdialog}. In the context of Task-Oriented (TO) dialogue, proactive dialogue definitions include non-collaborative dialogue as well as enriched TO dialogue. In non-collaborative dialogue, the agent and user have opposing objectives. Some examples include persuasion \cite{wang-etal-2019-persuasion, wu-etal-2021-alternating}, negotiation \cite{he-etal-2018-decoupling}, and deception-based dialogue \cite{santhanam-etal-2020-learning}. Enriched TO dialogue shares some similarities with our task. However, while enriched TO dialogue focuses on enhancing conversational naturalness through additional information, our goal is to sustain ISD. Rather than prioritizing naturalness, we aim to encourage user engagement by introducing new information (either directly or through a FQ) that prompts the user to continue the conversation.

With regard to ISD specifically, response proactivity largely revolves around generating clarifying questions and eliciting user preferences \cite{deng2023survey}. Clarifying question generation aims to resolve ambiguity in user queries to provide the user with the requested information \cite{aliannejadi-etal-2021-building}. Approaches include retrieval and ranking-based frameworks \cite{10.1145/3331184.3331265}, reinforcement learning with clarification utility rewards \cite{10.1145/3366423.3380126}, and multi-step frameworks predicting the need for a clarifying question before generating one \cite{aliannejadi-etal-2021-building, guo2021abgcoqa}. Some methods also combine clarifying questions and conversational QA in multi-turn context \cite{deng-etal-2022-pacific, guo2021abgcoqa}. User preference elicitation involves proactively reveals the user's interests for better recommendations \cite{10.1145/3269206.3271776}. This task is often treated as a decision-making problem often tackled with reinforcement learning \cite{10.1145/3269206.3271776, deng2021unified, jaques2019way}. Unlike earlier definitions, we do not concentrate on specific proactive ISD aspects like clarifying question generation or user preference elicitation. Instead, we solely focus on enhancing proactivity by providing relevant information. Moreover, we evaluate the proactiveness of each individual response separately, rather than considering the entire conversation.

\noindent\textbf{LLM-based ISD} In recent years, LLMs have emerged as leading models in language generation tasks, demonstrating state-of-the-art performance. In ISD, recent methods utilize LLMs through in-context learning or supervised fine-tuning. In-context learning refers to learning a new task during inference with a few prompt examples. Approaches leveraging few-shot \cite{li2023fewshot, chada2021fewshotqa} and CoT \cite{yoran-etal-2023-answering, sultan2024structured} prompts have been employed in this context. LLMs are also often trained on dialogue contexts alongside task instructions, which is known as instruction tuning, to enhance zero-shot performance. In the context of dialogue, LLMs such as Flan-T5 \cite{chung2022scaling}, InstructGPT \cite{ouyang2022training}, and InstructDial \cite{gupta2022instructdial} were explicitly trained on dialogue data for chat applications. Likewise, instruction-tuning has also been applied to improve the accuracy and informativeness of conversational QA responses \cite{jiang2024instructiontuned, razumovskaia2024textitdial}. 
These methods excel at achieving the primary aim of ISD to address user queries. However, as highlighted in Section~\ref{sec:intro}, they tend to produce reactive responses that do not proactively engage the user.

\begin{figure}
    \centering
    \includegraphics[scale=0.25]{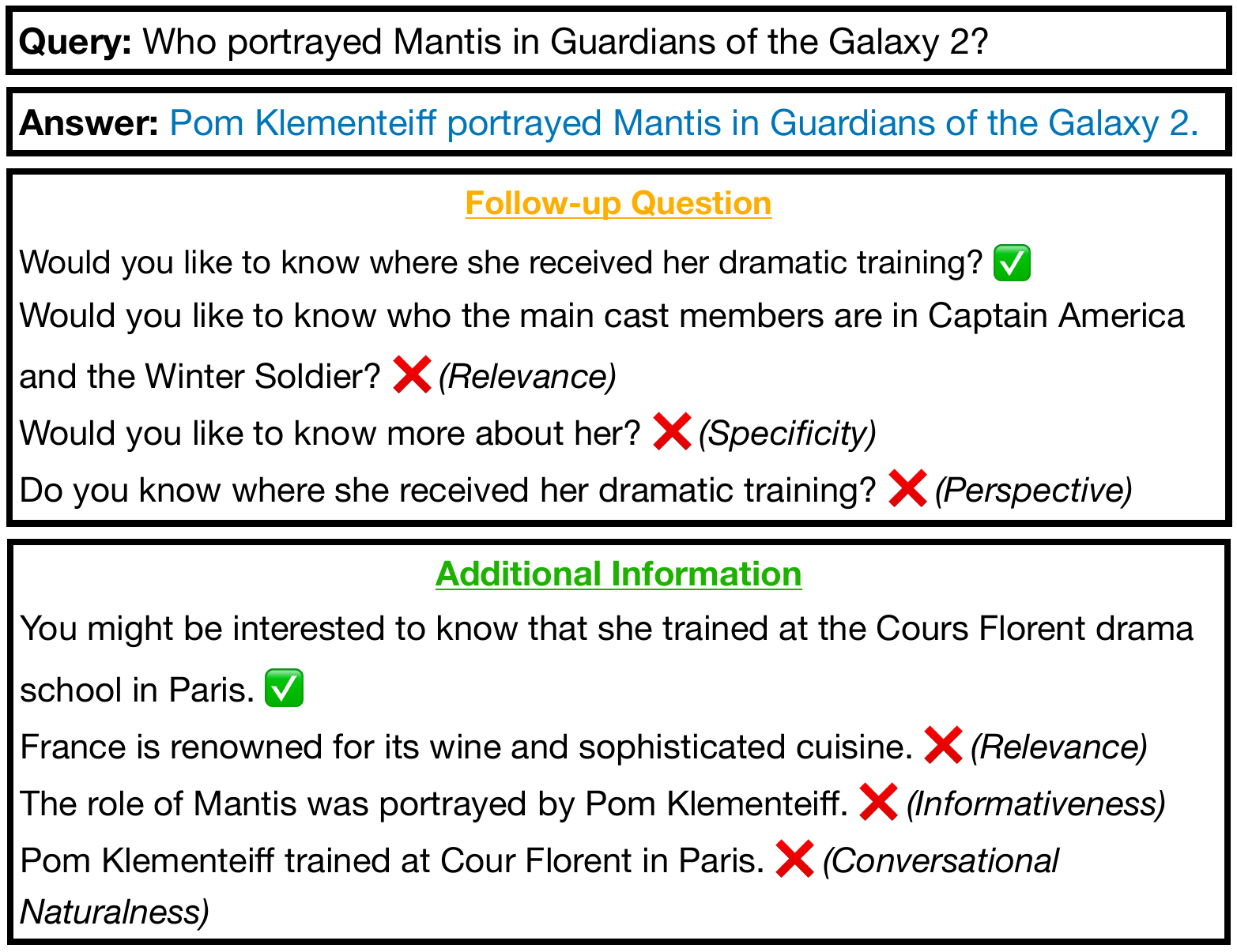}
    \caption{Examples of FQs and AI. Proactive elements that are accepted or unaccepted are symbolized by a green checkmark or a red "X" respectively. The criteria for deeming each proactive element as unacceptable is specified adjacent to the corresponding red "X".}
    \label{fig:critera_example}
\end{figure}

\section{Problem Definition}
\label{sec:problem}
We propose a new proactive response definition for ISD that consists of two components: an Answer and a Proactive Element. The Answer directly addresses the user's query, while the Proactive Element actively engages the user by providing related information. The proactive element enriches the user's understanding and can spark further interest, prompting them to further engage the conversation to find out more. We further classify the Proactive Element into two main categories: Additional Information (AI) and Follow-up Questions (FQs).

AI refers to any knowledge not explicitly requested in the user's query or mentioned in the answer, but that could be of interest to the user. The provision of high-quality AI enriches the conversation by increasing its informativeness, and encouraging the user to continue the interaction. To determine if an AI qualifies, the following criteria must be met:
\begin{enumerate}
    \item \emph{Relevance}. The AI should be relevant to the user's query.
    \item \emph{Informativeness}. The AI should provide substantial supplementary details beyond the original Answer. It should not be simply a rephrased version of the Answer.
    \item \emph{Naturalness}. The AI should be natural in a spoken conversational context. It should be introduced in a conversational manner and avoid excessive verbosity.
\end{enumerate}
It's important to note that LLMs often have a tendency to include excessive details in a single response, which can hinder naturalness, particularly in spoken context.
Our goal is to incorporate AI in a concise and engaging manner that encourages the user to continue the interaction.

A FQ asks if the user is interested in a specific piece of additional relevant information related to their initial query. The information itself is not explicitly provided in the FQ. By asking appropriate FQs, we can extend the conversation beyond the initial turn. The criteria for a FQ are defined as follows:
\begin{enumerate}
    \item \emph{Relevance}. The FQ should relate to knowledge relevant to the user's query.
    \item \emph{Specificity}.  The FQ should be as specific as possible, referring to a particular piece of information rather than making a broad inquiry. Specific FQs lead to more informative and satisfying interactions.
    \item \emph{Perspective}. The FQ should not request information from the user. It should focus on assisting and informing the user, avoiding information seeking.
\end{enumerate}
Figure 2 presents examples of responses that do not meet the previously mentioned criteria.


Unlike prior work in ISD, our definition focuses specifically on response proactivity rather than factual accuracy. Therefore, we do not include criteria related to information accuracy or ground responses on external knowledge sources. There factors are often used to prevent hallucination and ensure factual correctness. 
\section{Proactive Response Evaluation}
In this section, we propose several automatic metrics to quantify the proactivity of a response. A reliable automatic metric would enable objective and cost-effective evaluation, ultimately enhancing the reproducibility of our work.

\subsection{Baseline Metrics}
We introduce two baseline metrics: a prompt-based metric and a classification-based metric. The prompt-based metric, ranging from 0 to 1, is obtained by prompting an LLM to assess the proactiveness of responses based on our definition. The classification-based metric is calculated using two language models, each evaluating responses as valid or invalid for each Proactive Element type, according to our definition. More details are provided in Appendix \ref{app:baseline-metrics}.

\begin{figure*}
    \centering
    \includegraphics[scale=0.24]{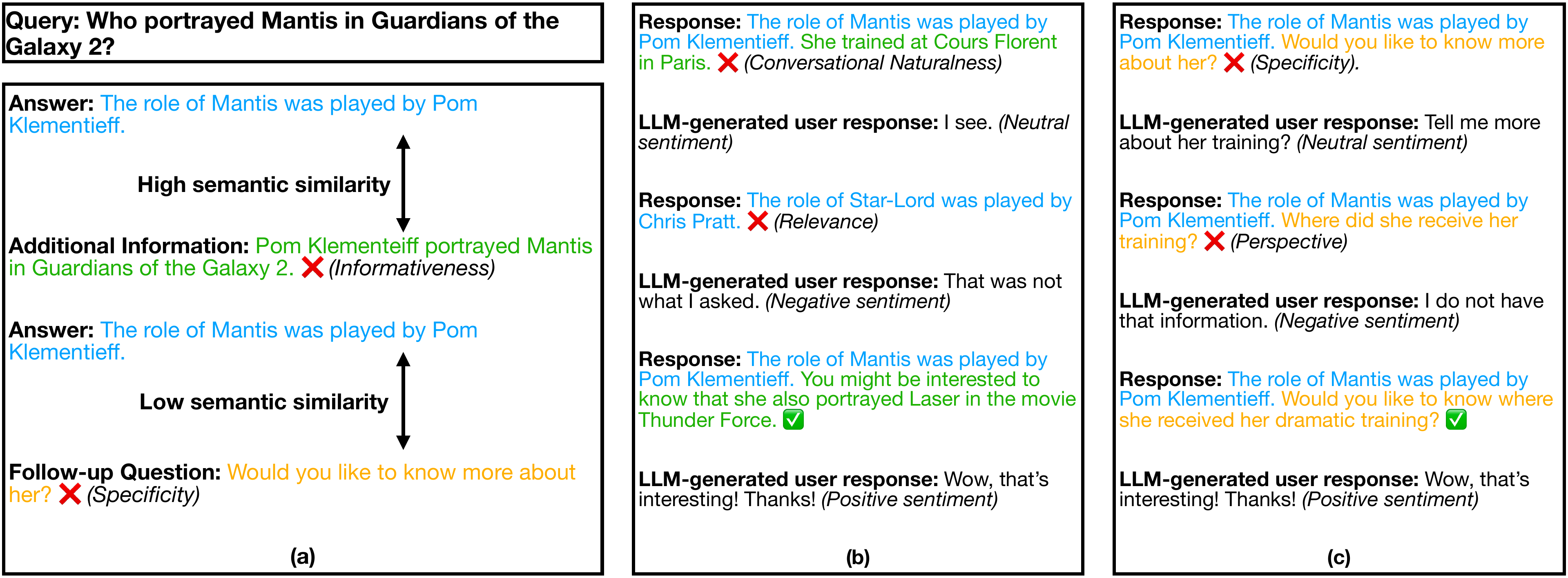}
    \caption{(a) Illustration of low and high semantic similarities in low quality AI and FQ respectively. (b) Samples of LLM-generated user responses for AI. (c) Samples of LLM-generated user responses for FQ.}
    \label{fig:metric_example}
\end{figure*}

\subsection{Proposed Metrics}
\label{sec:proposed_metrics}
The baseline scores often lack interpretability. They do not provide specific information about which criteria a response violates. Therefore, we propose two additional metrics which evaluate the responses based on the criteria defined in Section~\ref{sec:problem}.

\noindent\textbf{Semantic similarity-based} We design a metric based on semantic similarity to evaluate the \emph{Relevance} of a proactive response, as well as the \emph{Specificity} and \emph{Informativeness} of the FQ and AI respectively.

The respective semantic scores for the FQ and AI are computed as follows:
\begin{itemize}
\item FQ: $\alpha*BS(Q, R) + (1 - \alpha)\bar{BS}(R)$
\item AI: $\alpha*BS(Q, R) + (1 - \alpha)(1 - \bar{BS}(R))$
\end{itemize}
where $Q$ and $R$ denote the input query and generated response respectively. $BS(\cdot)$ refers to the BERTScore, and $\bar{BS}(res) = \frac{1}{n}\sum_{i,j | i \in n, j \in n}BS(r_{i}, r_{j})$, the mean pair-wise semantic similarity. $\alpha$ is a hyperparameter introduced to control the distribution between both terms. In our implementation, the BertScore is computed using the deberta-base-v3 embeddings.

It should also be highlighted that a completely irrelevant or incoherent proactive element would likely result in a lower semantic similarity score compared compared to a generic but related response. This difference is primarily due to the first term in the equations, which involves the BertScore calculation between the query and the response. An entirely irrelevant response would achieve a very low BertScore, whereas a generic but relevant response would obtain a relatively higher score. Conssequently, after appropriately adjusting $\alpha$, the semantic score for a proactive response containing irrelevant elements would be significantly low.

\noindent\textbf{User Simulation-based} We also propose a user simulation-based metric to quantify the quality of the Proactive Element based on \emph{Relevance} and \emph{Conversational Naturalness} of the AI, as well as the \emph{Specificity} and \emph{Perspective} of the FQ. This involves prompting an LLM to generate a simulated user turn in response to a given proactive system response, and then measuring the sentiment of the LLM-generated user response. After analyzing our initial responses, we found that users often react positively when we provide proactive responses paired with custom FQs or seamlessly integrated AI. This approach frequently elicits enthusiastic acknowledgments such as 'Yes, thank you!', 'Wow! That's interesting.', or 'That would be great. Thanks!', contributing to a LLM-generated user response with significantly positive sentiment. Conversely, subpar proactive responses that include generic FQs or conversationally unnatural AI tend to elicit replies with comparatively neutral sentiment. Furthermore, FQs with the wrong \emph{Perspective} (requesting information from the user) generally lead to more detailed responses containing the requested information, often resulting in a neutral sentiment. Naturally, responses that do not address the user's query will typically elicit responses with negative sentiment. Samples of generated responses and the corresponding LLM-generated user responses for AI and FQ are provided in Figure~\ref{fig:metric_example}(b) and \ref{fig:metric_example}(c) respectively. 

To obtain the user-simulation score, we prompt the LLM (with a temperature value of $t=0.5$) $n$ times to generate $n$ LLM-generated user responses. We then calculate the positive sentiment of each LLM-generated user response and take the average. The model used to determine positive sentiment is a fine-tuned RoBERTa pretrained language model \cite{camachocollados2022tweetnlp}. Any arbitrary LLM can used to generate the simulated user responses. This process is summarized in Algo \ref{alg:user-simul}.

\section{Corpus Construction}
To create our proactive dialogue corpus, we utilize the Natural Questions Question Answer (NQQA) dataset \cite{kwiatkowski-etal-2019-natural}. Each sample in this dataset includes a query, a short answer, and a long answer. The short answer provides the response to the query, while the long answer contains some relevant information. We selected the NQQA corpus because the query and short answer format resembles a typical single-turn conversation between a human and an ISD agent. However, since the short answer in the NQQA corpus consists of only a single entity, it needed to be modified for conversational naturalness.

\subsection{Annotation}
To achieve this, we engaged crowdworkers via Amazon Mechanical Turk (AMT) to modify the short answer to make it sound more like a natural response in a conversation, and to formulate the Proactive Element. AMT instructions are provided in Appendix \ref{app:amt-instruct}. These two components were concatenated to form the final proactive response. This process allowed us to construct a proactive dialogue corpus that could be used for training and evaluating proactive ISD agents.

\noindent\textbf{Answer} The Answer component is obtained by enhancing the short answer found in the NQQA corpus. This short answer, which is the direct answer to the user's query, is modified to ensure conversational naturalness. The crowdworkers were given instructions to integrate the short response, often a single verb or noun, into a coherent and comprehensive sentence that effectively addresses the user's query in a conversational style. For example, for the query in Figure~\ref{fig:dial_example} and \ref{fig:critera_example}, the short response ('Pom Klementieff') resulted in the following sentence: 'The actress who portrayed Mantis in Guardians of the Galaxy is Pom Klementieff'.


\noindent\textbf{Proactive Element} To obatin the Proactive Element (FQ or AI), crowdworkers were provided the long answer for reference. This simplified the task and ensured the accuracy of the Proactive Elements. For FQs, crowdworkers were instructed to create inquiries that assessed whether the user desired a particular piece of information from the long answer. They were encouraged to make their questions as specific as possible, focusing on particular details rather than general inquiries. For AI, crowdworkers were told to identify a single piece of information not already present in the initial answer and rephrase it to sound more natural in a conversational context. Before annotation, we filtered the NQQA dataset based on query length and long answer length. This ensured the clarity of the query and guaranteed that there was sufficient information from which the crowdworkers can formulate either a FQ or AI.

\subsection{Corpus Features and Statistics}
Based on the approach described above, we extracted 1000 samples and collect 2,000 proactive dialogue samples (1,000 for each Proactive Element) for our proactive response corpus. Each sample in our corpus constitutes a single-turn dialogue consisting of a user query and a proactive response. After obtaining the annotations, we manually validated each response to ensure fluency and correct any spelling or grammatical errors. The number of samples and average query length are identical for both Proactive Elements as a single query is used to obtain two proactive responses, one for each Proactive Element. Some basic corpus statistics are provided in Appendix \ref{app:corpus}.


\section{Proactive Response Generation}
In this section, we describe the in-context learning and instruction-tuning approaches we employed for proactive response generation.

\subsection{In-context Learning}
In-context learning involves explicitly providing demonstrations of the task at hand to the model as part of a prompt. In this section, we describe three in-context learning prompts we utilize for proactive response generation: the direct prompt, 3-step CoT prompt, and 3-in-1 CoT prompt. For our experiments, we implemented 0-shot, 1-shot, and 3-shot variants of these three prompts. Prompt templates are provided in Appendix \ref{app:prompt-templates}.

\noindent\textbf{Direct Prompt} This approach involves direct prompting the LLM to generate answers with the task description and demonstrations of query-proactive response pairs.

\noindent\textbf{3-step Chain-of-Thought (CoT) Prompt} We introduce a 3-step CoT prompting approach designed to effectively generate proactive responses. Our approach involves systematically decomposing the proactive response generation task into three distinct subtasks, each addressed by an independent prompt. This entails three separate inferences. The output from each prompt is used as input for the subsequent prompt. The three prompts corresponding to the three subtasks are as follows:
\begin{itemize}
    \item[$P_{1}:$] Query answering: In this step, the LLM is prompted to generate the precise answer to the user's query.
    \item[$P_{2}:$] Related information generation: Building upon the answer generated in $P_{1}$, the LLM is directed to identify a specific piece of related information that was not present in the initial answer.
    \item[$P_{3}:$] Proactive Element generation: For the FQ, the LLM is prompted to formulate an inquiry to ask the user if they would like to receive the information generated in $P_{2}$. Alternatively, for the AI, the LLM is prompted to rephrase the content produced in $P_{2}$ in a manner that reflects a scenario where the information is being offered to the user.
\end{itemize}
The final proactive response $R$ is obtained by combining the output of $P_{1}$ and the output of $P_{3}$, i.e., $R = LLM(P_{1}) + LLM(P_{3})$, where $+$ refers to the concatenate operation. We conduct simple post processing (rule-based removal of escape characters as well as excess spacing) on the output of each prompt to ensure the quality of the input to the subsequent prompt.

In the 1-shot and 3-shot versions, demonstration examples were not provided to $P_{1}$ as $P_{1}$ achieved good performance in the 0-shot setting. Additionally, since the reference information from which the response is based on is not readily available in our corpus, $P_{2}$ and $P_{3}$ would entail manually deriving the reference information for few-shot prompting.

\noindent\textbf{3-in-1 Chain-of-Thought (CoT) Prompt} A drawback of the previous approach is the necessity for three distinct model inferences, leading to increased latency during generation. To address this, we attempt to consolidate all three prompts into a single 3-in-1 prompt. This unified prompt provides explicit instructions to the LLM to follow the exact same process as before in a step-by-step manner, encompassing all three subtasks within a single inference. We also implement a 0-shot, 1-shot, and 3-shot version of this prompt. Unlike the 3-stop CoT prompt, no manual derivation of specific information is required. Only the query and response, which are readily available, is required.

\noindent\textbf{Demonstration Selection} We also perform demonstration selection using metrics outlined in Section~\ref{sec:proposed_metrics}.
Specifically, we identify the top-$k$ and bottom-$k$ responses (for a $k$-shot prompt) using the following criteria: (1) the user-simulation score, (2) the semantic similarity score, and (3) the sum of both scores. Generally, we observe that using the sum of both scores results in the generation of high-quality responses that achieve high user-simulation and semantic similarity scores. Full results are provided in Appendix \ref{app:demo-selection}.

\subsection{Instruction Tuning}
We also instruction tuned an LLM via QLoRA \cite{dettmers2023qlora} to generate proactive responses. Leveraging our proposed corpus, we conducted instruction tuning on two distinct tasks corresponding to the generation of proactive responses with either a FQ or AI. We utilized 1000 proactive responses (500 from each proactive element). 

\section{Experiments}

\begin{table}[]
\centering
\scalebox{0.7}{
\begin{tabular}{@{}lcc@{}}
\toprule
                & \multicolumn{1}{c}{FQ} & \multicolumn{1}{c}{AI} \\ \midrule
Prompt-based & -0.072                                                                             & 0.163          \\
Classification-based    & 0.188                                                                             & 0.492     \\
User Simulation-based & 0.256                                                                             & 0.331                                                                                 \\
Semantic Similarity-based & \textbf{0.462}                                                                             & \textbf{0.575}                                                                                                                  \\ \bottomrule
\end{tabular}}
\caption{Point Biserial correlations between our proposed user-simulation, semantic similarity, prompt-based, and classification-based scores and human annotation.}
\label{tbl:corr}
\end{table}

\begin{table*}[!ht]
\centering
\scalebox{0.6}{
\begin{tblr}{
  row{odd} = {c},
  row{2} = {c},
  row{4} = {c},
  row{6} = {c},
  row{8} = {c},
  row{10} = {c},
  cell{1}{3} = {c=4}{},
  cell{1}{7} = {c=4}{},
  cell{3}{1} = {r=3}{},
  cell{6}{1} = {r=3}{},
  cell{9}{1} = {r=3}{},
  cell{12}{3} = {c},
  cell{12}{4} = {c},
  cell{12}{5} = {c},
  cell{12}{6} = {c},
  cell{12}{7} = {c},
  cell{12}{8} = {c},
  cell{12}{9} = {c},
  cell{12}{10} = {c},
  hline{1,3,6,9,12-14} = {-}{},
  hline{2} = {3-10}{}}
                &        & FQ      &                    &                   &                     & AI  &                    &                   &                     \\
                &        & {Classification} & {User Simulation} & {Semantic Similarity} & {Num Token} & {Classification} & {User Simulation} & {Semantic Similarity} & {Num Token} \\
Direct          & 0-shot & 0.73                   & 0.45              & 0.32             & 20.35              & 0.52                   & 0.49              & 0.28             & 28.53              \\
                & 1-shot & 0.92                   & 0.51              & 0.51             & 30.55              & 0.74                   & 0.51              & 0.33             & 33.67              \\
                & 3-shot & 0.92                   & 0.52              & 0.59             & 28.90              & 0.79                   & 0.52              & 0.37             & 30.07              \\
{3-step\\CoT}   & 0-shot & 0.88                   & 0.51              & 0.59             & 32.45              & 0.86                   & 0.49              & 0.31             & 38.65              \\
                & 1-shot & 0.93                   & 0.53              & 0.61             & 34.73              & 0.81                   & 0.52              & 0.35             & 37.18              \\
                & 3-shot & \textbf{0.95}                   & 0.53              & 0.62             & 31.79              & 0.90                   & 0.54              & 0.38             & 39.28               \\
{3-in-1\\CoT  } & 0-shot & 0.68                   & 0.46              & 0.39             & 23.93              & 0.44                   & 0.51              & 0.26             & 26.09              \\
                & 1-shot & 0.90                   & 0.52              & 0.60             & 29.65              & 0.93                   & 0.56              & 0.40             & 38.50              \\
                & 3-shot & 0.92                   & 0.51              & 0.60             & 34.86              & 0.95                   & \textbf{0.63}              & \textbf{0.41}             & 34.64              \\

SFT & & 0.94          & \textbf{0.54}     & \textbf{0.64}    & 28.24     & \textbf{0.96}  & 0.55 & \textbf{0.41} & 32.10  \\

Human           &        & 0.96                   & 0.55              & 0.63             & 28.33              & 0.97                   & 0.67              & 0.43             & 36.18              
\end{tblr}}
\caption{Classification, Semantic similarity, user-simulation scores, and average token length when direct prompting, 3-step prompting, 3-in-1 prompting, and instruction-tuning is applied to Falcon-40b-instruct. The highest score for each metric, other than the scores for the human generated responses, is \textbf{bolded}.}
\label{tbl:results}
\end{table*}

\noindent\textbf{Instruction Tuning}{Implementation} In our experiments, we utilize the 40b instruction-tuned Falcon LLM \cite{refinedweb} and the 13b StableVicuna LLM\cite{vicuna2023}. Results attained using StableVicuna are provided in the Appendix \ref{app:vicuna}. We utilize a temperature value of 0.2 for all generations.  For each Proactive Element, we split our proactive dialogue corpus into two distinct sets: a 500-sample training set and a 500-sample test set. We select demonstration examples for our prompts from the training sets, and then evaluate them on the test set. We instruction-tune the LLM on the training sets for both the FQs and AI concurrently. The instructions used are identical to the direct prompt.

\noindent\textbf{Metric Correlations} Table~\ref{tbl:corr} shows the Point Biserial correlations between our new metrics and human annotations, calculated from a dataset of 500 positive samples from our corpus and 500 negative samples generated by prompting a LLM for subpar proactive responses that lack a proactive element, feature low-quality proactive element or are completely irrelevant with respect to the user's input.

The prompt-based baseline yields low correlation scores, highlighting its limitations as a metric. Conversely, the classification-based baseline achieve better, though inconsistent, correlations with human evaluations. Specifically, correlations for AI are higher than those for FQs. This difference arises because negative samples for AI, which mostly violate the \emph{Informativeness} criteria, are simpler for the model to detect compared to the nuanced, generic responses that characterize negative samples for FQs, which violate the \emph{Specificity} criteria.  Future research could involve improving the correlations through further prompt engineering or by enriching the training dataset with more varied negative examples.

The proposed semantic and sentiment scores clearly outperform both baselines. The semantic metric, encompassing \emph{Relevance}, \emph{Informativeness} (AI), and \emph{Specificity} (FQs), achieves the highest correlation scores. This aligns with expectations, as many negative responses lack the required \emph{Informativeness} and \emph{Relevance}. Conversely, the sentiment score focuses on \emph{Perspective} and \emph{Conversational Naturalness}, which are less common in negative samples. Therefore, we recommend using both metrics together to effectively evaluate response proactiveness, covering the criteria outlined in Section~\ref{sec:problem} comprehensively.

\noindent\textbf{In-Context Learning} Scores attained by the direct, 3-step CoT, and 3-in-1 CoT prompts on Falcon-40b-instruct are shown in Table~\ref{tbl:results}. A key finding is that the 3-step CoT prompt generally enhances 0-shot performance, addressing the general lack of proactive element seen in responses in the 0-shot direct and 3-in-1 CoT prompts, which generate fewer tokens in the 0-shot setting. The 3-step prompt resolves this by ensuring the final proactive response includes FQs or AI by concatenating outputs from the 1st and 3rd prompts.

It is also evident that the 3-step CoT prompt surpasses both the 3-in-1 CoT and direct prompts when it comes to the FQ. Conversely, for AI, the 3-in-1 CoT prompt outperforms both the 3-step CoT and direct prompts. This could be attributed to the inherent difficulty in generating high-quality FQs for the LLM, which generally excels at generating informative responses. Consequently, the FQ task benefits more from the 3-step CoT prompt since it breaks down the task into three simpler components.


\noindent\textbf{Instruction Tuning} Table~\ref{tbl:results} also includes results for the instruction-tuned Falcon-40b-instruct, which produced responses similar to the 3-shot variants of the 3-step and 3-in-1 CoT prompts for FQs and AI, respectively. These responses strictly adhere to the structure outlined in Section~\ref{sec:problem}. Compared to prompted responses, there are fewer instances of missing Answers or Proactive Elements. Instead, lower-quality responses lacked \emph{Specificity} (FQs) or \emph{Conversational Naturalness} (AI).

\section{Multi-turn Setting}

To demonstrate the efficacy of our approach in the multi-turn conversations, we sampled 50 test cases from our dataset and interactions between a simulated user and an agent using Falcon-40b-instruct. We used 3-step and 3-in-1 CoT prompts with modifications to produce proactive responses, detailed in Appendix \ref{app:mutiturn}. 

After conducting 50 simulations, we discovered that when the agent includes AI or FQ, the user is significantly more inclined to continue interacting with the agent. In contrast, responses lacking this proactive element usually consist of the agent merely acknowledging the information provided, naturally ending the conversation (Table~\ref{tbl:res_standard}). From the 50 simulations conducted, we found that approximately 94\% of conversations ended after just one turn. In contrast, only 22\% and 34\% of interactions with the agent generating proactive responses with FQ and AI respectively ended after a single turn. On average, users continued the conversation for 3.9 turns with the FQ agent and 3.2 turns with the AI agent before ending the conversation naturally. For the FQ, the simulated user naturally requests the agent to provide the information suggested by the agent, further sustaining the interaction and improving the informativeness of the whole conversation (Table~\ref{tbl:res_followup}). For AI, the AI provided by the agent would tend to elicit more involved responses from the  user rather than a cursory acknowledgement (Table~\ref{tbl:res_addinfo}) as well as encourage the user to inquire further about the AI provided by the agent. 

However, both proactive elements displayed a tendency to repeat the proactive element from earlier in the conversation. We hypothesize that this issue could potentially be alleviated by improving quality of the LLM. To confirm our hypothesis, we repeat the experiment using GPT-4 instead of Falcon-40b-instruct for the Assistant. The sample conversations demonstrate that GPT-4 effectively minimizes such repetitions across up to four dialogue turns (Table~\ref{tbl:res_chatgpt}). In our experiments, we apply our prompts at every conversational turn. However, in real-world ISD, not every turn would warrant a proactive response. Future work could constitute introducing an approach to detect if a proactive response is appropriate.

\section{Conclusion}

In this work, we propose a novel response-level definition of ISD proactivity. Per our definition, a proactive response includes both an Answer and a Proactive Element (FQ or AI). We compiled a dataset consisting of 2000 single-turn dialogues, and introduced a novel 3-step CoT and 3-in-1 CoT prompt that outperforms standard few-shot prompts in generating proactive responses. Future work could entail exploring finer-grained proactive elements or employing reward modelling and Reinforcement Learning with Human Feedback (RLHF) for fine-tuning. Expanding the current corpus to the multi-turn scenarios could also facilitate further research to improve in-context learning or supervised fine-tuning performance. Existing conversation-level metrics in ISD could also be enhanced to account for response-level proactivity. The performance of different LLMs on our task can also be explored.

\section{Limitations}
Firstly, the effectiveness of the generation approaches proposed are highly dependent on the LLMs that underpin them. Hence, different LLMs may display inherent biases or produce unforeseen outputs, resulting in lower quality response sets. Secondly, there are limitations based on the computational resources available. We do not have the capability to conduct in-context learning or instruction tuning experiments with larger or more recent LLMs. Future work could entail the evaluating the zero-shot performance of these LLMs on our proposed task. Thirdly, in this work, we do not assert that our prompt template is the optimal choice for proactive response generation. Our direct, CoT and 3-step CoT prompt templates are intended to form a baseline for researchers to improve upon. Additional work could entail additional, more deliberate prompt engineering.

\section{Ethics Statement}

We recruited annotators ("Turkers") through Amazon Mechanical Turk to build our dataset. Each Turker received detailed information about the Human Intelligence Task (HIT), including task descriptions, requirements and compensation, before agreeing to participate. They were free to withdraw from the task at any time for any reason. Each Turker was compensated at the rate of 0.20USD per HIT, and each HIT took an average of 55.6 seconds (12.90USD per hour).
\bibliography{anthology,custom}
\bibliographystyle{acl_natbib}

\clearpage
\appendix
\section{Appendix}

\subsection{Demonstration Selection}
\label{app:demo-selection}
\begin{table*}
\centering
\scalebox{0.65}{
\begin{tblr}{
  cells = {c},
  cell{1}{3} = {c=4}{},
  cell{1}{7} = {c=4}{},
  cell{3}{1} = {r=3}{},
  cell{6}{1} = {r=3}{},
  hline{1,3,6,9-10} = {-}{},
  hline{2} = {3-10}{},
}
         &           & Follow-up Question &           &          &           & Additional Information &           &          &           \\
         &           & Classification     & User Simulation & Semantic Similarity & Num Token & Classification         & User Simulation & Semantic Similarity & Num Token \\
Top-1    & Semantic  & 0.83               & 0.48      & \textbf{0.54}     & 31.22     & 0.72                   & 0.47      & \textbf{0.36}     & 32.98     \\
         & Sentiment & 0.82               & \textbf{0.53}      & 0.50     & 32.53     & \textbf{0.76}                   & \textbf{0.53}      & 0.34     & 34.51     \\
         & Sum       & \textbf{0.87}               & 0.51      & 0.52     & 30.94     & 0.74                   & 0.50      & 0.34     & 35.68     \\
Bottom-1 & Semantic  & 0.80               & 0.49      & 0.42     & 28.53     & 0.69                   & 0.45      & 0.30     & 30.51     \\
         & Sentiment & 0.79               & 0.44      & 0.46     & 27.22     & 0.65                   & 0.42      & 0.31     & 31.22     \\
         & Sum       & 0.76               & 0.44      & 0.45     & 26.38     & 0.66                   & 0.43      & 0.31     & 33.27     \\
Random   &           & 0.81               & 0.51      & 0.51     & 30.55     & 0.74                   & 0.51      & 0.33     & 33.67     
\end{tblr}}
\caption{Classification, user-simulation, semantic similarity scores, and average token length when demonstration selection is applied to 1-shot direct prompting on Falcon-40b-instruct. The highest score for each metric is \textbf{bolded}.}
\label{tbl:demo-1shot}
\end{table*}

\begin{table*}
\centering
\scalebox{0.65}{
\begin{tblr}{
  cells = {c},
  cell{1}{3} = {c=4}{},
  cell{1}{7} = {c=4}{},
  cell{3}{1} = {r=3}{},
  cell{6}{1} = {r=3}{},
  hline{1,3,6,9-10} = {-}{},
  hline{2} = {3-10}{},
}
         &           & Follow-up Question      &                    &                   &                     & Additional Information  &                    &                   &                     \\
         &           & {Classification} & {User Simulation} & {Semantic Similarity} & {Num Token} & {Classification} & {User Simulation} & {Semantic Similarity} & {Num Token} \\
Top-3    & Semantic  & 0.92                   & 0.54              & \textbf{0.62}             & 31.32              & 0.90                   & 0.53              & \textbf{0.41}             & 32.01              \\
         & Sentiment & 0.92                   & \textbf{0.56}              & 0.59             & 31.57              & \textbf{0.91}                   & \textbf{0.58}              & 0.38             & 32.67              \\
         & Sum       & \textbf{0.94}                   & 0.56              & 0.60             & 30.73              & \textbf{0.91}                   & 0.54              & 0.40             & 36.62              \\
Bottom-3 & Semantic  & 0.84                   & 0.47              & 0.53             & 25.63              & 0.50                   & 0.41              & 0.28             & 23.62              \\
         & Sentiment & 0.80                   & 0.45              & 0.56             & 27.51              & 0.76                   & 0.44              & 0.32             & 34.09              \\
         & Sum       & 0.73                   & 0.44              & 0.55             & 28.43              & 0.75                   & 0.47              & 0.36             & 26.61              \\
Random   &           & 0.92                   & 0.52              & 0.58             & 28.33              & 0.79                   & 0.52              & 0.36             & 30.07              
\end{tblr}}
\caption{Classification, user-simulation, semantic similarity scores, and average token length when demonstration selection is applied to 3-shot direct prompting on Falcon-40b-instruct. The highest score for each metric is \textbf{bolded}.}
\label{tbl:demo-3shot}
\end{table*}

\begin{table*}
\centering
\scalebox{0.65}{
\begin{tblr}{
  cells = {c},
  cell{1}{3} = {c=4}{},
  cell{1}{7} = {c=4}{},
  cell{3}{1} = {r=3}{},
  cell{6}{1} = {r=3}{},
  hline{1,3,6,9-10} = {-}{},
  hline{2} = {3-10}{},
}
         &           & Follow-up Question &           &          &           & Additional Information &               &               &           \\
         &           & Classification     & User Simulation & Semantic Similarity & Num Token & Classification         & User Simulation     & Semantic Similarity      & Num Token \\
Top-5    & Semantic  & 0.92               & 0.53      & \textbf{0.63}     & 30.91     & 0.90                   & 0.56          & \textbf{0.43} & 33.75     \\
         & Sentiment & 0.91               & \textbf{0.56}      & 0.60     & 31.82     & 0.92                   & \textbf{0.59}          & 0.36          & 30.33     \\
         & Sum       & \textbf{0.93}               & 0.55      & 0.61     & 29.72     & \textbf{0.94}                   & 0.57          & 0.41          & 32.14     \\
Bottom-5 & Semantic  & 0.86               & 0.43      & 0.56     & 25.37     & 0.57                   & 0.45          & 0.26          & 37.25     \\
         & Sentiment & 0.90               & 0.45      & 0.59     & 27.46     & 0.61                   & 0.47          & 0.32          & 25.81     \\
         & Sum       & 0.87               & 0.42      & 0.58      & 28.32     & 0.56                  & 0.42          & 0.34          & 24.76     \\
Random   &           & 0.92               & 0.53      & 0.61     & 29.29     & 0.83                   & 0.55          & 0.35          & 31.75     
\end{tblr}}
\caption{Classification, user-simulation, semantic similarity scores, and average token length when demonstration selection is applied to 5-shot direct prompting on Falcon-40b-instruct. The highest score for each metric is \textbf{bolded}.}
\label{tbl:demo-5shot}
\end{table*}

\begin{table*}
\centering
\scalebox{0.65}{
\begin{tblr}{
  row{odd} = {c},
  row{2} = {c},
  row{4} = {c},
  row{6} = {c},
  row{8} = {c},
  row{10} = {c},
  cell{1}{3} = {c=4}{},
  cell{1}{7} = {c=4}{},
  cell{3}{1} = {r=3}{},
  cell{6}{1} = {r=3}{},
  cell{9}{1} = {r=3}{},
  cell{12}{3} = {c},
  cell{12}{4} = {c},
  cell{12}{5} = {c},
  cell{12}{6} = {c},
  cell{12}{7} = {c},
  cell{12}{8} = {c},
  cell{12}{9} = {c},
  cell{12}{10} = {c},
  hline{1,3,6,9,12-14} = {-}{},
  hline{2} = {3-10}{},
}
              &        & Follow-up Question &                 &                     &           & Additional Information &                 &                     &           \\
              &        & Classification     & User Simulation & Semantic Similarity & Num Token & Classification         & User Simulation & Semantic Similarity & Num Token \\
Direct        & 0-shot & 0.33               & 0.35            & 0.36                & 22.39     & 0.61                   & 0.29            & 0.29                & 26.83     \\
              & 1-shot & 0.65               & 0.37            & 0.57                & 25.97     & 0.67                   & 0.31            & 0.33                & 28.74     \\
              & 3-shot & 0.86               & 0.39            & 0.60                & 27.85     & 0.69                   & 0.35            & 0.31                & 29.16     \\
{3-step\\CoT} & 0-shot & 0.82               & 0.41            & 0.58                & 33.90     & 0.75                   & 0.36            & 0.31                & 30.24     \\
              & 1-shot & 0.86               & 0.43            & 0.59                & 35.75     & 0.78                   & 0.40            & 0.36                & 29.46     \\
              & 3-shot & \textbf{0.92}      & \textbf{0.47}   & \textbf{0.61}       & 28.42     & 0.82                   & 0.42            & 0.37                & 29.25     \\
{3-in-1\\CoT} & 0-shot & 0.72               & 0.42            & 0.41                & 27.51     & 0.68                   & 0.37            & 0.32                & 25.21     \\
              & 1-shot & 0.82               & 0.41            & 0.57                & 26.45     & 0.85                   & 0.42            & 0.39                & 36.42     \\
              & 3-shot & 0.91               & 0.43            & 0.59                & 25.94     & \textbf{0.92}          & \textbf{0.44}   & \textbf{0.40}       & 38.51     \\
Human         &        & 0.96               & 0.55            & 0.63                & 28.33     & 0.97                   & 0.67            & 0.43                & 36.18     
\end{tblr}}
\caption{Classification, semantic similarity, user simulation scores, and average token length when direct prompting, 3-step CoT prompting, 3-in-1 CoT prompting, and instruction-tuning (SFT) is applied to StableVicuna. The highest score for each metric, other than the scores for the human generated responses, is \textbf{bolded}.}
\label{tbl:vicuna}
\end{table*}

\begin{figure*}
    \centering
    \includegraphics[scale=0.4]{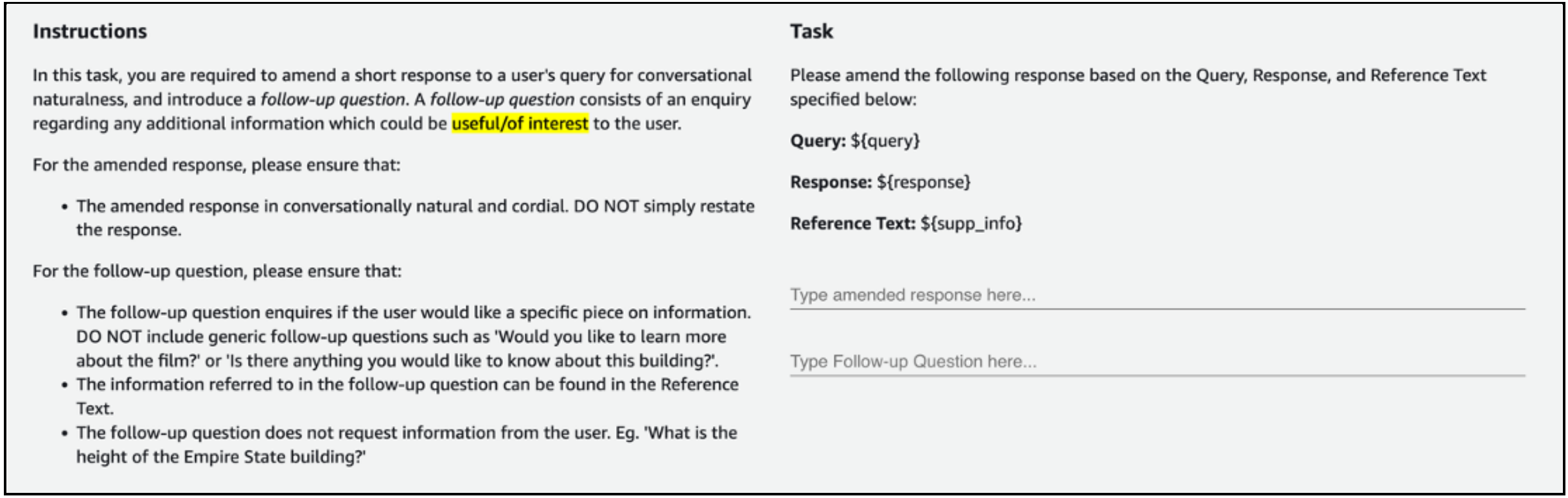}
    \caption{AMT instructions for the FQ.}
    \label{fig:amt-followup}
\end{figure*}
\begin{figure*}
    \centering
    \includegraphics[scale=0.4]{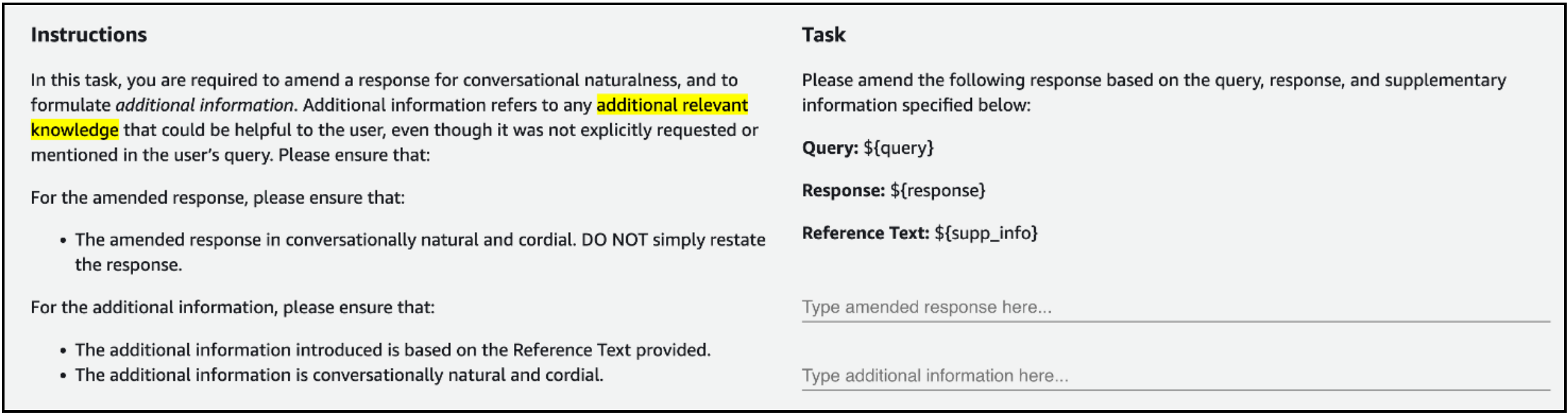}
    \caption{AMT instructions for the AI.}
    \label{fig:amt-addinfo}
\end{figure*}

Results for 1-shot, 3-shot and 5-shot demonstration selection are presented in Table \ref{tbl:demo-1shot}, \ref{tbl:demo-3shot}, and \ref{tbl:demo-5shot} respectively. 

Generally, the results attained align closely with our expectations. When we select demonstration examples using sentiment or semantic metrics as criteria, the resulting responses tend to achieve higher scores in the user simulation and semantic similarity scores respectively. For example, with regard to the FQ, selecting the bottom-1, 3, or 5 examples based on the semantic score would result in relatively generic FQs, which are reflected in the low semantic similarity scores. Similarly, for the AI, selecting the top-1, 3, or 5 examples based on sentiment score would result in responses with conversationally natural AI and high user simulation scores. 

Also, while there is a slight decrease in semantic similarity score when bottom examples are selected based on semantic similarity for the AI, this drop is minimal. Especially when compared to the drop in user simulation score brought about by selecting the bottom examples based on sentiment for the FQ. This is primarily due to the fact that the responses in our dataset largely meet the criteria of \emph{Informativeness} for the AI, leading to an overall high semantic similarity score. On the other hand, there is a relatively  larger variance in terms of quality with regard to \emph{Specificity} for the FQ (eg. 'Would you like to know more about Pom Guardians of the Galaxy 2?' vs 'Would you like to know who portrayed the character of Peter Quill in Guardians of the Galaxy 2?'). 

Additionally, it can be observed that while there is a relatively significant increase in performance between 1 and 3-shot prompts, the 3-shot and 5-shot prompts generally achieve comparable performance. It should also be noted that when we select demonstration examples based on the sum of the sentiment and semantic metrics, the generated responses exhibit balanced improvements across all criteria.

\subsection{StableVicuna}
\label{app:vicuna}
The scores attained when direct, 3-step CoT, and 3-in-1 CoT prompting are applied to StableVicuna are provided in Table \ref{tbl:vicuna}. 

Generally, the trends observed in the results and responses attained via Falcon-40b-instruct can be observed in the case of StableVicuna. The 3-step CoT and 3-in-1 CoT prompts generally improve on 0-shot performance. Also, for the FQ, the performance of the 3-step CoT prompt exceeds both the 3-in-1 CoT and direct prompts. For AI, the 3-in-1 CoT prompt achieves better performance compared to both the 3-step CoT and direct prompts.

In addition, with the exception of the semantic similarity score, Falcon-40b-instruct generally attains higher scores across all metrics. When it comes to the semantic similarity, responses generated by StableVicuna and Falcon-40b-instruct attained comparable scores. This suggests that, in terms of providing AI, StableVicuna's responses exhibit a relatively lower level of \emph{Naturalness} compared to Falcon-40b-instruct. In other words, the AI in the responses tend to be introduced in a relatively abrupt fashion as opposed to a conversationally natural manner. For the FQ, StableVicuna's responses exhibit a comparatively lower level of specificity when compared to those generated by Falcon-40b-instruct. The FQs from StableVicuna more often refer to general, broad areas which would likely require further specification from the user.

\subsection{AMT Instruction}
\label{app:amt-instruct}
Throughout the data collection process, several pilot tests were conducted in order to refine the instructions provided to the turkers via AMT. The final instructions and interface utilized during data collection are provided in Fig \ref{fig:amt-followup} and \ref{fig:amt-addinfo} respectively. For both the FQ and AI, three turkers were engaged at a rate of 0.20USD per task (or HIT). 

\subsubsection{Answer}

Firstly, turkers were instructed to amend a reference response for conversational naturalness to attain the Answer component. Initially, the turkers were instructed to input the Answer and the Proactive Element in a single input field. However, during the initial pilot tests, we found that numerous turkers simply input the reference response provided as is, without any amendment. The reference response corresponds to the short answer from the Natural Questions QA corpus, which consists of a single entity (eg. 'Pom Kleimentieff', '4th of July', or 'United States of America'). This negatively impacts the naturalness of the overall proactive response. We found that this issue can be addressed by breaking down the task into two distinct components with separate instructions and input fields. One for amending the reference response, and another for formulating the proactive element. Positive and negative examples were also included to place further emphasis on the importance of amending the reference response.

\subsubsection{Follow-up Question}

For the FQ, turkers were told to formulate a FQ that references a specific piece of information in the reference text provided. The reference text corresponds to the long answer in the Questions QA corpus. The initial pilot tests revealed a strong tendency for turkers to input extremely short and generic questions (eg. 'Would you like to know more?', 'Are you interested in learning more?'). Hence, the final instructions explicitly highlight the importance of ensuring that the questions are as specific as possible, in addition to emphasizing that the question should not request any information from the user. Positive and negative examples were provided for the user's reference. 

\subsubsection{Additional Information}

For the AI, the turkers were instructed to formulate a additional relevant information based on the reference text provided. Providing turkers with the reference text serves to ensure the factuality of the AI formulated. For this HIT, the main issue found in the initial pilot tests centered on conversational naturalness. Turkers were formulating AI which resembled factual statements as opposed to information introduced in a conversational manner ('Chris Pratt portrayed Star Lord in Guardians of the Galaxy.' vs. 'Did you know that Chris Pratt played the role of Star Lord in Guardians of the Galaxy?'). To mitigate this issue, turkers were explicitly instructed to ensure that the AI was conversationally natural and cordial. Additionally, positive and negative examples were similarly provided for the user's reference.

\subsection{Corpus Statistics}
\label{app:corpus}
To provide a broad overview of the corpus, we compute the average query length, response length, as well as Proactive Element length for each Proactive Element type. The derived statistics are provided in Table \ref{tbl:corpus}.
\begin{table}[]
\centering
\scalebox{0.7}{
\begin{tabular}{@{}lcc@{}}
\toprule
                                                                                          & \begin{tabular}[c]{@{}c@{}}Follow-up\\ Question\end{tabular} & \begin{tabular}[c]{@{}c@{}}Additional\\ Information\end{tabular} \\ \midrule
Number of Samples                                                                & 1000                                                                  & 1000                                                                      \\
\begin{tabular}[c]{@{}l@{}}Average Tokens per\\ Query\end{tabular}               & 11.054                                                                & 11.054                                                                    \\
\begin{tabular}[c]{@{}l@{}}Average Tokens per \\ Proactive Response\end{tabular} & 28.293                                                                & 35.607                                                                    \\
\begin{tabular}[c]{@{}l@{}}Average Tokens per \\ Proactive Element\end{tabular}  & 14.711                                                                & 22.736                                                                    \\ \bottomrule
\end{tabular}}
\caption{Proactive dialogue corpus statistics.}
\label{tbl:corpus}
\end{table}

\subsection{Prompt Templates}
\label{app:prompt-templates}

\begin{figure*}
    \centering
    \includegraphics[scale=0.4]{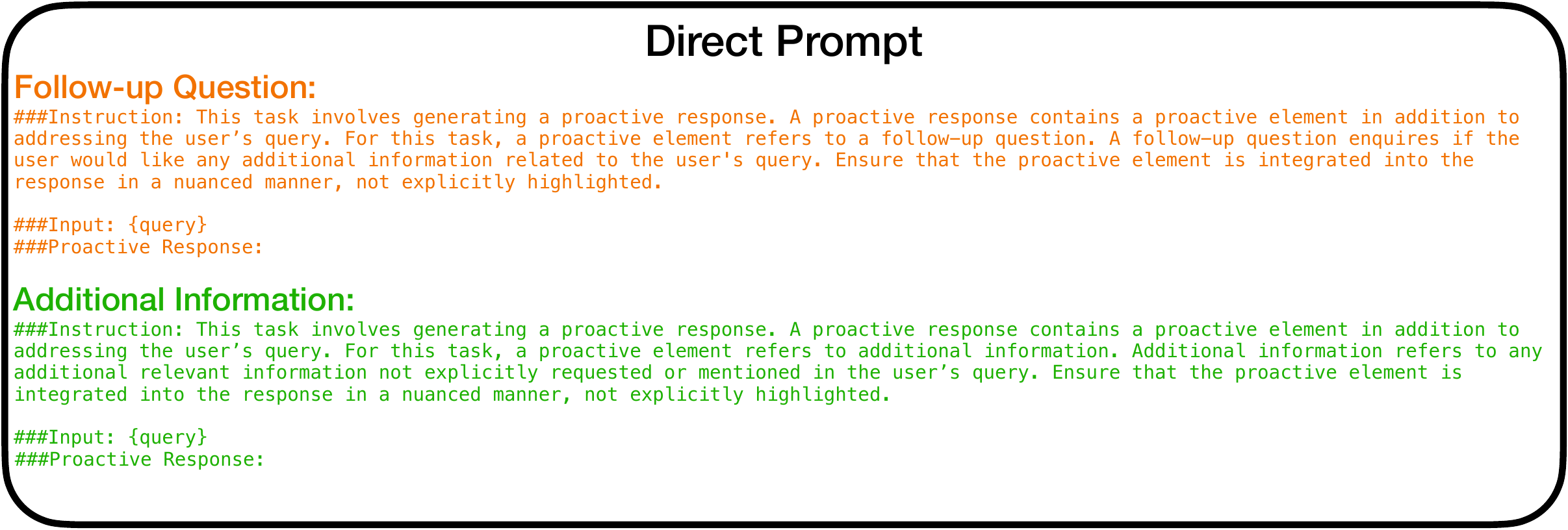}
    \caption{Direct prompt template.}
    \label{fig:direct-template}
\end{figure*}

\begin{figure*}
    \centering
    \includegraphics[scale=0.4]{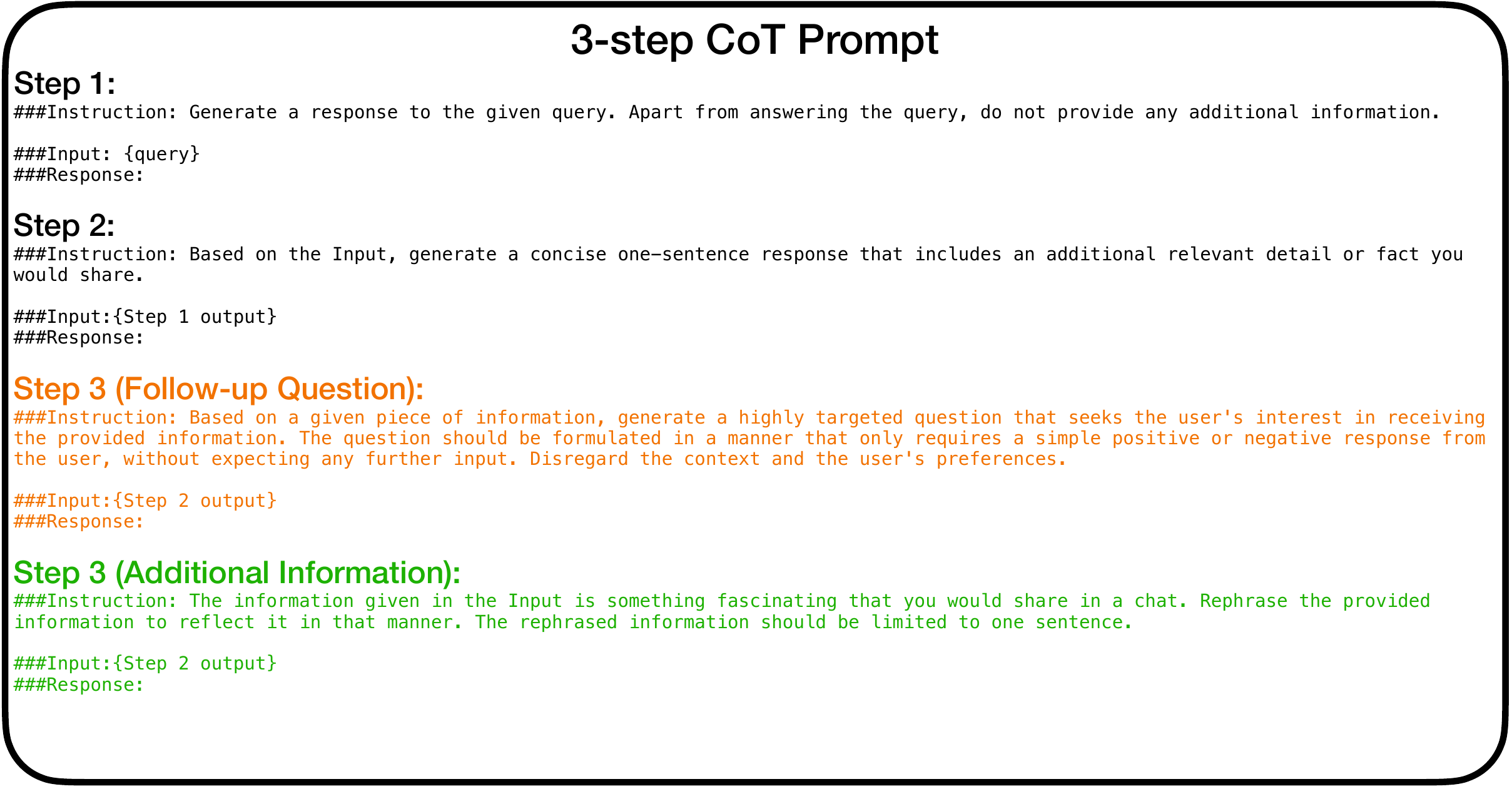}
    \caption{3-step CoT prompt template.}
    \label{fig:3step-template}
\end{figure*}

\begin{figure*}
    \centering
    \includegraphics[scale=0.4]{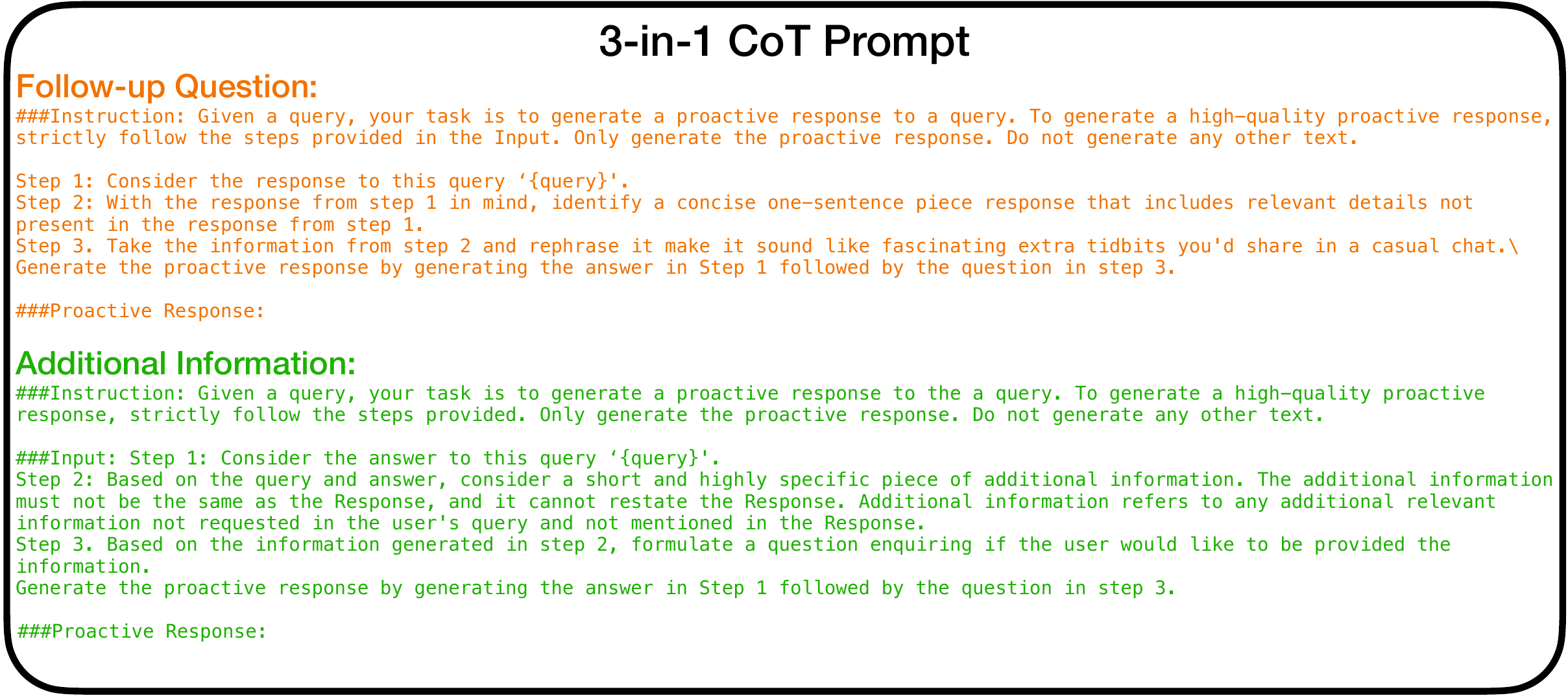}
    \caption{3-in-1 CoT prompt template.}
    \label{fig:3in1-template}
\end{figure*}

The prompt templates for the direct, 3-step, and 3-in-1 CoT prompts are provided in Figure \ref{fig:direct-template}, \ref{fig:3step-template} and \ref{fig:3in1-template} respectively. Prompts specific to the FQ are in \textcolor{orange}{orange}, and prompts specific to the AI are in \textcolor{green}{green}.


\subsection{Baseline Metrics}
\label{app:baseline-metrics}
Alongside the semantic similarity and user-simulation scores, we introduce two straightforward baseline metrics: a prompt-based metric and a classification-based metric. 
\subsubsection{Prompt-based}
Our prompt-based approach is based on \citet{jain-etal-2023-multi}, where a few-shot prompt is used to generate a score to quantify various dimensions of quality in text summarization. In our context, we similarly leverage a LLM to generate a score (ranging from 0 to 1) that indicates the proactiveness of the response based on our definition. We craft two prompts, one for each Proactive Element. Each prompt includes the task description and annotated proactive response-score pairs. A response scoring 1.0 is proactive and meets all criteria in Section 2, while a 0.0 score indicates a lack of Proactive Element and failure to address the user's query. Responses meeting one or two criteria are scored 0.25 and 0.75, respectively. We use Falcon-40b-instruct\cite{refinedweb} with a temperature of 0 for deterministic responses. We then parse the numeric string to attain the Prompt-based score. 

For the prompt, we experimented with several different demonstration examples. The five examples selected adhered to the following format: one perfect proactive response that fulfilled every criteria, one response that violated one of the criterion, two responses which violated two different criteria, and one response which violated all three criteria. We observed that as long as the examples provided followed this format, varying the examples and their quantity did not significantly affect correlation with human annotation.

\subsubsection{Classification-based}
We introduce a model-based metric for measuring response proactiveness. To achieve this, we finetune two language models, one for each Proactive Element type, to classify responses as either valid or invalid in accordance with our definition. For fine-tuning, we utilize a small annotated dataset of 700 samples (with a 500/100/100 split) which consists of 59\% valid and 41\% invalid responses. For this task, we utilize the DeBERTa-V3-large model from Huggingface \cite{he2021debertav3}, which attained accuracy of 0.80 and 0.84 on the AI and FQ respectively. Subsequently, the final model score is attained by extracting the positive logit value during inference. 
\subsubsection{User-Simulation Score Algorithm}
An algorithm detailing the step-by-step procedure to compute the user-simulation score is provided in Algorithm \ref{alg:user-simul}.
\begin{algorithm}
\caption{User-simulation score computation.}\label{alg:cap}
\begin{algorithmic}
\Require $n$, $t$,$R$
\State $S \gets 0$ 
\While{$n \neq 0$}
    \State $R_{LLM} \gets LLM_{t}(R)$ \Comment{Attain LLM-generated user response.}
    \State $S_{temp} = Sentiment_{pos}(R_{LLM})$ \Comment{Compute positive sentiment}
    \State $S \gets S + S_{temp}$ 
    \State $n \gets n - 1$
\EndWhile
\State $score \gets \frac{1}{n} S$ \Comment{Compute average}
\State \textbf{return} $score$
\end{algorithmic}
\label{alg:user-simul}
\end{algorithm}

\subsection{Response Samples}
\label{app:response-samples}
Samples of responses generated via direct prompting, 3-step CoT prompting, 3-in-1 CoT prompting, and instruction-tuning (SFT) for the FQ and AI are provided in Table \ref{tbl:samples-followup} and \ref{tbl:samples-addinfo} respectively. For the FQ, the Answer component is missing from the response generated by the 0-shot direct prompt. The responses generated by the direct 1-shot prompt and the 3-in-1 CoT prompt are relatively lacking in terms of \emph{Specificity}. Responses generated by the remaining prompts generally fulfill all criteria outlined in Section 3. For the AI, the 0-shot direct and 3-in-1 prompt are missing the Proactive Element, and both the 1-shot direct and 1-shot 3-step prompts generated responses that lacked in \emph{Naturalness}. The remaining prompts largely satisfied all the criteria described in Section 3.

\begin{table*}
\centering
\begin{tblr}{
  row{2} = {c},
  cell{1}{1} = {c},
  cell{1}{2} = {c},
  cell{2}{2} = {c=2}{},
  cell{3}{1} = {c},
  cell{3}{2} = {c},
  cell{4}{1} = {r=3}{c},
  cell{4}{2} = {c},
  cell{5}{2} = {c},
  cell{6}{2} = {c},
  cell{7}{1} = {r=3}{c},
  cell{7}{2} = {c},
  cell{8}{2} = {c},
  cell{9}{2} = {c},
  cell{10}{1} = {r=3}{c},
  cell{10}{2} = {c},
  cell{11}{2} = {c},
  cell{12}{2} = {c},
  cell{13}{1} = {c},
  hline{1-4,7,10,13-14} = {-}{},
}
User Query    &           & How far do honey bees travel from their hive?                                                                                                                                                                         \\
              & Responses &                                                                                                                                                                                                                       \\
Human         &           & {Honey bees travel around four miles from their hive, but the exact distance \\depends on a number of factors. Would you like to learn more about the foraging \\habits of honey bees?}                                  \\
Direct        & 0-shot    & Would you like to know the distance honey bees travel from their hive?                                                                                                                                                \\
              & 1-shot    & {Honey bees typically travel up to 3 miles from their hive. Would you like\\~to know more about the behavior of honey bees?}                                                                                          \\
              & 3-shot    & {Honey bees can travel up to 3 miles from their hive. Would you like to know\\~how they navigate and communicate with each other during their flight?}                                                                \\
{3-step\\CoT} & 0-shot    & {Honey bees typically travel up to 3 miles from their hive to collect nectar and \\pollen. Would you like to learn more about the navigation skills of honey bees?}                                                   \\
              & 1-shot    & {Honey bees typically travel up to 3 miles from their hive in search of nectar and\\pollen. Would you like to know how honey bees navigate?} \\
              & 3-shot    & {Honey bees can travel up to 3 miles from their hive in search of nectar and pollen. \\Would you like to know more about the amazing sense of smell of honey bees?}                                                   \\
{3-in-1\\CoT} & 0-shot    & {Honey bees can travel up to 3 miles from their hive in search of food and water. \\Would you like to know more about the fascinating behavior of honey bees?}                                                        \\
              & 1-shot    & {Honey bees can travel up to 3 miles from their hive. Would you like to know\\~more about the bees flight range?}                                                                                                     \\
              & 3-shot    & {Honey bees can travel up to 3 miles from their hive. Would you like to know \\how they navigate and communicate with each other during their travels?}                                                               \\
SFT           &           & {Honey bees travel up to three miles from their hive. Do you want to know more \\about the reasons behind the particular behavior of honey bees?}                                                                     
\end{tblr}
\caption{Samples of proactive responses (FQ) generated by Falcon-40b-instruct via direct, 3-step, and 3-in-1 CoT prompting as well as the instruction tuning.}
\label{tbl:samples-followup}
\end{table*}

\begin{table*}
\centering
\begin{tblr}{
  row{2} = {c},
  cell{1}{1} = {c},
  cell{1}{2} = {c},
  cell{2}{2} = {c=2}{},
  cell{3}{1} = {c},
  cell{3}{2} = {c},
  cell{4}{1} = {r=3}{c},
  cell{4}{2} = {c},
  cell{5}{2} = {c},
  cell{6}{2} = {c},
  cell{7}{1} = {r=3}{c},
  cell{7}{2} = {c},
  cell{8}{2} = {c},
  cell{9}{2} = {c},
  cell{10}{1} = {r=3}{c},
  cell{10}{2} = {c},
  cell{11}{2} = {c},
  cell{12}{2} = {c},
  cell{13}{1} = {c},
  hline{1-4,7,10,13-14} = {-}{},
}
User Query    &           & Who does Michael Scott end up with in The Office?                                                                                                                                                         \\
              & Responses &                                                                                                                                                                                                           \\
Human         &           & {Michael Scott ends up with Holly Flax. It might interest you to know that the\\character Michael Scott is portrayed by Steve Carell.}                                                                   \\
Direct        & 0-shot    & Michael Scott ends up with Holly Flax in the Office.                                                                                                                                                      \\
              & 1-shot    & {Michael Scott ends up with Holly Flax in the Office. They have a long-distance \\relationship that eventually leads to marriage.}                                                                        \\
              & 3-shot    & {Michael Scott ends up with Holly Flax in the Office. Here's a fun fact, they got \\married in the season 7 finale.}                                                                                      \\
{3-step\\CoT} & 0-shot    & {Michael Scott ends up with Holly Flax in the Office. Michael Scott's\\relationship with Holly Flax in the Office is a great example of how love can\\blossom in unexpected places.}                      \\
              & 1-shot    & {Michael Scott ends up with Holly Flax in the finale of The Office. That's what\\she said is a running joke in The Office, often used by Michael Scott.}                                                 \\
              & 3-shot    & {Michael Scott ends up with Holly Flax in the Office.  You might be surprised to\\find out that Michael Scott and Holly Flax started a romantic relationship only\\in the later seasons of The Office.} \\
{3-in-1\\CoT} & 0-shot    & Michael Scott ends up with Holly Flax in the end.                                                                                                                                                         \\
              & 1-shot    & {Michael Scott ends up with Holly Flax in the final season of The Office. You might be\\interested to know that they got married and moved to Colorado.}                                                                           \\
              & 3-shot    & {Michael Scott ends up with Holly Flax in the end. Did you know that they had\\four beautiful children together?}                                                                         \\
SFT           &           & {Michael Scott ends up with Holly Flax. It may interest you to know that he proposes\\to her towards the end of the seventh season.}                           
\end{tblr}
\caption{Samples of proactive responses (additional information) generated by Falcon-40b-instruct via direct, 3-step, and 3-in-1 CoT prompting as well as the instruction tuning.}
\label{tbl:samples-addinfo}
\end{table*}

\subsection{Multi-turn Conversation Samples}
\label{app:mutiturn}
To demonstrate the efficacy of our in-context learning approach in the multi-turn setting, we sample 50 test cases from our dataset and initiate a conversation using each case as the initial user query. We utilize Falcon-40b-instruct to simulate a human user interacting with an ISD agent. For the human user simulation, the LLM is simply fed the dialogue context and instructed to generate the appropriate response. 

The proactive ISD agent also consist of Falcon-40b-instruct prompted via the 3-step prompt (FQ) and 3-in-1 prompt (AI). In addition, we also implement an ISD agent prompted to simply respond to the user's query, without a proactive element. The 3-in-1 prompt template is largely identical to the one used in single-turn experiments, with one key difference: instead of a single query, a dialogue context is provided. Slight textual edits (substituting the word `query' with `dialogue context') were made in the prompt instructions to incorporate the dialogue context, thereby ensuring the LLM is aware that the input is a dialogue context context rather than solely a single query. Similarly, for the first prompt $P_{1}$ in the 3-step prompt, the dialogue context is provided rather than a single query, and the prompt instructions were edited accordingly. For the second prompt $P_{2}$, the outputs generated from $P_{1}$ during the conversation are concatenated and utilized in $P_{2}$ instead. No other modifications were made to the prompt templates.

Tables \ref{tbl:res_standard}, \ref{tbl:res_followup}, and \ref{tbl:res_addinfo} display sample conversations where the ISD agent is instructed to respond directly to the user’s query, generate a proactive response with a FQs, and generate a proactive response with additional questions, respectively. Table \ref{tbl:res_chatgpt} features conversations where GPT-4 was used to generate the Assistant's responses. 
\begin{table*}
\centering
\begin{tblr}{
  hline{1,4,7} = {-}{},
}
User:      & Who is the actor who played William Beech in Goodnight Mr Tom?   \\
Assistant: & Nicholas Farrell played William Beech in Goodnight Mr Tom.       \\
User:      & Thank you.                                                       \\
User:      & How big is the screen at Minute Maid park?                       \\
Assistant: & The screen at Minute Maid park is 30 feet tall and 55 feet wide. \\
User:      & Thanks for the information.                                      
\end{tblr}
\caption{Conversation between a simulated human user (User) and an ISD agent(Assistant). After the Assistant provides the required information, the User simply acknowledges, thereby ending the conversation.}
\label{tbl:res_standard}
\end{table*}

\begin{table*}
\centering
\begin{tblr}{
  hline{1,11,19} = {-}{},
}
User:      & Who is the actor who played William Beech in Goodnight Mr Tom?                                                                                                                                                                                                                                                                                                           \\
Assistant: & {The actor who played William Beech in the movie 'Goodnight Mr. Tom' is Oliver Ford\\Davies. Would you like to know more about the actor Oliver Ford Davies and his career?}                                                                                                                                                                                             \\
User:      & Yes, I would like to know more about Oliver Ford Davies and his career.                                                                                                                                                                                                                                                                                                  \\
Assistant: & {Oliver Ford Davies is an English actor who has appeared in numerous films, television\\shows, and stage productions. He is known for his roles in Goodnight Mr. Tom, The Lion\\in Winter, and The Return of the Native. He has also appeared in several Shakespeare plays.\\Would you like to know which three plays are considered to be Shakespeare's greatest work?} \\
User:      & Yes, I would like to know which three plays are considered to be Shakespeare’s greatest works.                                                                                                                                                                                                                                                                           \\
Assistant: & {The three plays that are considered to be Shakespeare's greatest works are Hamlet, Macbeth,\\and King Lear. Would you like to know what is the name of the main character in the play\\King Lear?}                                                                                                                                                                      \\
User:      & Yes, I would like to know what is the name of the main character in the play King Lear.                                                                                                                                                                                                                                                                                  \\
Assistant: & {The name of the main character in the play King Lear is King Lear himself. Would you like\\to know the name of the actor who played King Lear in the film?}                                                                                                                                                                                                             \\
User:      & Yes, I would like to know the name of the actor who played King Lear in the film.                                                                                                                                                                                                                                                                                        \\
Assistant: & {The actor who played King Lear in the film is Sir Laurence Olivier. Would you like to know\\the name of the king in the play King Lear?}                                                                                                                                                                                                                                \\
User:      & How big is the screen at Minute Maid park?                                                                                                                                                                                                                                                                                                                               \\
Assistant: & {The screen at Minute Maid Park is 54 feet high and 124 feet wide. \textbf{Would you like to know}\\\textbf{the capacity of Minute Maid Park?}}                                                                                                                                                                                                                           \\
User:      & Yes, please. Tell me the capacity of Minute Maid Park.                                                                                                                                                                                                                                                                                                                   \\
Assistant: & {The capacity of Minute Maid Park is approximately 40,000 people. Would you like to know\\in which city the park is located?}                                                                                                                                                                                                                                            \\
User:      & Yes, I would like to know in which city the park is located.                                                                                                                                                                                                                                                                                                             \\
Assistant: & {The Minute Maid Park is located in Houston, Texas. Would you like to know the name of the\\baseball team that plays in Minute Maid Park?}                                                                                                                                                                                                                               \\
User:      & Yes, please.                                                                                                                                                                                                                                                                                                                                                             \\
Assistant: & {The name of the baseball team that plays in Minute Maid Park is the Houston Astros. \textbf{Would}\\\textbf{you like to know the capacity of Minute Maid Park?}}                                                                                                                                                                                                         
\end{tblr}
\caption{Multi-turn conversation between a simulated human user (User) and an ISD agent(Assistant). Falcon-40b-instruct is utilized to simulate both the User and Assistant. The 3-step FQ prompt is used to generate the Assistant's responses. The \textbf{bold} text in the second conversation denotes the repeated AI.}
\label{tbl:res_followup}
\end{table*}

\begin{table*}
\centering
\begin{tblr}{
  hline{1,9,17} = {-}{},
}
User:      & Who is the actor who played William Beech in Goodnight Mr Tom?                                                                                                                                                                                                                                                                                                                                                                               \\
Assistant: & {The actor who played William Beech in Goodnight Mr. Tom is Nicholas Farrell. Did you know\\that Nicholas Farrell also played the role of Captain Charles Lightoller in the Titanic movie?~}                                                                                                                                                                                                                                                 \\
User:      & No, I did not know that. That's interesting. What other movies has Nicholas Farrell been in?~                                                                                                                                                                                                                                                                                                                                                 \\
Assistant: & {Nicholas Farrell has been in many movies,including The Madness of King George, The Jewel\\in the Crown, and The English Patient. He has also appeared in several TV shows, such as\\The Crown and The Night Manager.}                                                                                                                                                                                                                       \\
User:      & Wow, he has been in a lot of movies and TV shows. ~                                                                                                                                                                                                                                                                                                                                                                                          \\
Assistant: & {He has played a variety of roles throughout his career including Cardinal Borgia in The Borgias.\\Did you know that he also appeared in several Shakespearean productions, including The Merchant\\of Venice and The Tempest.} \\
User:      & No, I did not know that. That's interesting.                                                                                                                                                                                                                                                                                                                                                                                                  \\
Assistant: & {Nicholas Farrell is an English actor known for his role in Chariots of Fire and The English Patient.\\It might also interest you to know that he also played the role of Ebenezer Scrooge in A Christmas\\Carol.}                                                                                                                                                                               \\
User:      & How big is the screen at Minute Maid park?                                                                                                                                                                                                                                                                                                                                                                                                   \\
Assistant: & {\textbf{The screen at Minute Maid Park is 54 feet high and 315 feet wide, making it one of the}\\\textbf{largest screens in the world.} Did you know that Minute Maid Park is the first retractable-roof\\stadium in the United States?~}                                                                                                                                                                                                   \\
User:      & {Wow, that's amazing! I didn't know that. What other unique features does Minute Maid Park\\have?~}                                                                                                                                                                                                                                                                                                                                             \\
Assistant: & {Minute Maid Park is also known for its unique train-themed features, including a locomotive\\shaped scoreboard and a train-shaped playground for kids. In addition, its also home to the Astros\\mascot, Orbit, who rides a train around the park during games.~}                                                                                                                                                                           \\
User:      & Wow, that's really interesting! I didn't know that Minute Maid Park had so many unique features.                                                                                                                                                                                                                                                                                                                                             \\
Assistant: & { Minute Maid Park also has a total capacity of approximately 40,000 people. \textbf{It might also interest}\\\textbf{you to know that the screen at Minute Maid park is 54 feet high and 315 feet long.}}                                                                                                                                                                                                                                        \\
User:      & {Wow, that's impressive! I didn't know that Minute Maid Park has such as large screen. I'm sure\\that the fans love it!~}                                                                                                                                                                                                                             \\
Assistant: & {Minute Maid Park also has a train that runs around the stadium and a swimming pool in the\\outfield. Did you know that the name of the baseball team that plays in Minute Maid Park\\is the Houston Astros?}                                                                                                                                                                                                                                
\end{tblr}
\caption{Multi-turn conversation between a simulated human user (User) and an ISD agent(Assistant). Falcon-40b-instruct is utilized to simulate both the User and Assistant. The 3-in-1 AI prompt is used to generate the Assistant's responses. The \textbf{bold} text in the second conversation denotes the repeated AI.}
\label{tbl:res_addinfo}
\end{table*}

\begin{table*}
\centering
\begin{tblr}{
  row{1} = {c},
  row{10} = {c},
  cell{1}{1} = {c=2}{},
  cell{10}{1} = {c=2}{},
  hline{1-2,11,19} = {-}{},
  hline{10} = {1}{l},
  hline{10} = {2}{r},
}
\textbf{Follow-up Question}     &                                                                                                                                                                                                                                                                           \\
User:                           & How big is the screen at Minute Maid park?                                                                                                                                                                                                           \\
Assistant:                      & {The screen at Minute Maid Park is approximately 54 feet high by 124 feet wide. Would you\\like to know which MLB team plays at Minute Maid Park?}                                                                                                                           \\
User:                           & Yes, I would like to know which MLB team plays at Minute Maid Park.                                                                                                                                                                                                       \\
Assistant:                      & {The Houston Astros play at Minute Maid Park. Would you like to know the specific feature\\of Minute Maid Park designed to handle the climate?}                                                                                                                              \\
User:                           & Sure, I'd love to know the feature designed to handle the climate.                                                                                                                                                                                                        \\
Assistant:                      & {Minute Maid Park features a retractable roof to handle the hot and humid climate of Houston,\\Texas. Would you like to know which year the Houston Astros won their first World Series title?}                                                                              \\
User:                           & Yes, please.                                                                                                                                                                                                                                                              \\
Assistant:                      & {The Houston Astros won their first World Series title in 2017. Would you like to know when Orbit\\was introduced as the team's mascot?}                                                                                                                                     \\
\textbf{Additional Information} &                                                                                                                                                                                                                                                                           \\
User:                           & How big is the screen at Minute Maid park?                                                                                                                                                                                                                                \\
Assistant:                      & {The screen at Minute Maid park is actually one of the largest in baseball, measuring over 55\\feet wide and 30 feet tall! Did you know that it's also one of the first retractable roof stadiums\\in the country?}                                                           \\
User:                           & That’s interesting! What other unique features does Minute Maid Park have?                                                                                                                                                                                                \\
Assistant:                      & {Minute Maid park also has one of the first retractable roof stadiums in the country. Hey, speaking\\of unique features, did you know it has a train on the outfield wall that moves whenever the Astros\\hit a home run?}                                                    \\
User:                           & Yes, I've heard about the train! It's an iconic feature of Minute Maid Park. What else?                                                                                                                                                                                   \\
Assistant:                      & {Minute Maid Park is also known for its unique, asymmetrical field dimensions. Did you know that\\these unconventional dimensions can affect the outcome of the games played there?}                                                                                         \\
User:                           & I did not know that. That’s interesting!                                                                                                                                                                                                                                  \\
Assistant                       & {Minute Maid Park also features a strikingly realistic artificial hill, known as Tal's Hill, which\\was in center field until 2016. Did you know that this hill was a tribute to classic ballparks with\\uneven terrain, adding an extra layer of challenge for outfielders?} 
\end{tblr}
\caption{Multi-turn conversation between a simulated human user (User) and an ISD agent(Assistant). GPT-4 is utilized to simulate the Assistant.}
\label{tbl:res_chatgpt}
\end{table*}

\clearpage

\end{document}